\documentclass{article}
\usepackage{spconf,amsmath,graphicx}
\usepackage{float}
\usepackage{subcaption}
\usepackage{eso-pic}
\graphicspath{{images}{images/}}

\title{Sequential IoT Data Augmentation using Generative Adversarial Networks}
\name{Maximilian Ernst Tschuchnig and Cornelia Ferner and Stefan Wegenkittl\thanks{Work on this publication was supported by the Information Technologies Institute of the University of Applied Sciences and the research project 860282 ERP4Cloud (EFRETop) managed by Manfred Mayr.}}
\address{Salzburg University of Applied Sciences\\
	Puch bei Hallein, Austria}
\begin{document}
%\ninept
%
\AddToShipoutPicture*{\small \sffamily\raisebox{1.2cm}{\hspace{1.8cm}978-1-5386-5541-2/18/\$31.00 ©2018 European Union}}
\maketitle
\begin{abstract}
Sequential data in industrial applications can be used to train and evaluate machine learning models (e.g. classifiers). Since gathering representative amounts of data is difficult and time consuming, there is an incentive to generate it from a small ground truth. Data augmentation is a common method to generate more data through a priori knowledge with one specific method, so called generative adversarial networks (GANs), enabling data generation from noise. This paper investigates the possibility of using GANs in order to augment sequential Internet of Things (IoT) data, with an example implementation that generates household energy consumption data with and without swimming pools. The results of the example implementation seem subjectively similar to the original data. Additionally to this subjective evaluation, the paper also introduces a quantitative evaluation technique for GANs if labels are provided. The positive results from the evaluation support the initial assumption that generating sequential data from a small ground truth is possible. This means that tedious data acquisition of sequential data can be shortened. In the future, the results of this paper may be included as a tool in machine learning, tackling the small data challenge.
\end{abstract}
\section{Motivation}
\label{sec:mot}

In the Internet of Things (IoT), where devices are connected and share information, sequential data, in the form of time-based sensor readings, are used in applications such as predictive maintenance \cite{Kanawaday2017} (e.g. in the form of malfunction detection), performance estimation or home automation. Being able to predict malfunctions can decrease maintenance costs and reduce upkeep times \cite{gubbi2013}, while home automation increases the control inhabitants have about their IoT devices \cite{Pavithra2015}. Performance estimation, on the other hand, can be used to predict outcomes, depending on changes of specific inputs, comparable to regression. Since ML and especially deep learning (DL) need substantial amounts of labelled, representative data to train its models, a tedious data acquisition phase is typically needed in order to utilize these applications \cite[p.~163]{goodfellow2016}. Especially when dealing with private or even sensitive data (e.g. home automation), as well as machines which have a rare occasion of malfunctions (predictive maintenance), collecting a significant amount of data conventionally is difficult. In some of these cases, data augmentation can be applied through the use of priori knowledge to effectively increase the starting data amount, minimizing acquisition delays. Looking at image data, basic approaches to data augmentation include patch extraction, or translation and rotation \cite{wang2017}.

Apart from classical data augmentation, deep neural networks can be used to generate data by training convolutional neural networks (CNNs), as proposed by Lecun et al. \cite{lecun1995}, to generate data from noise. These networks are trained by updating transposed convolutional kernels to generate realistic images from some input like noise, which are called "generators". To actually train the network in a meaningful way, another network, a so called "discriminator" or "critic" is constructed which tries to classify if an image is real or generated. The composition of these networks, so called generative adversarial networks (GANs), proposed by Goodfellow et al. \cite{goodfellow2014}, have had widespread success in the generation of image data but are rarely applied to sequential 1D data. This paper is going to investigate the possibility of generating 1D sequential data by first of all converting it into 2D, with the goal of exploiting locality (local features) in the original signal and a following GAN implementation.

\paragraph*{Related Work} Conventional data augmentation typically enriches the dataset through a priori knowledge. Since deep (representation) neural networks are able to learn distributed representations by themselves \cite[p.~4]{goodfellow2016}, they can be used to find representations in data and, in the context of GANs, apply this knowledge to construct new data from the previously learned representations. Therefore, a prior knowledge is no longer needed since GANs learn to apply it automatically \cite{goodfellow2014}. Similarly, autoencoders, an unsupervised representation learning model, learn representations from the original data $x$ to a reconstruction $P(Q(x))$ through an internal code $Q(x)$. Their target is to get $P(Q(x))$ to be similar to $x$ with the intermediate step $Q(x)$ having less neurons than $x$ and $P(Q(x))$ have features. Variational autoencoders (VAEs) add noise ($z$) to the internal code $Q(x|z)$, which, if reconstructed into some $P(Q(x|z))$, results in an augmented form of the original $x$ \cite{gregor2015}. Although VAEs are typically applied to image data, they can also be applied to sequential data if this data is converted using the process introduced in section \ref{sec:hecd}.

Another way of generating sequential data is to apply recurrent neural networks (RNNs) or long-short term memory networks (LSTMs). These networks take a sequence of a specific size as an input and generate data following the given sequence. More formally, a RNN, trained on a series of inputs $(x_1, x_2, ..., x_T)$, uses the outputs of the applied test data $(o_1, o_2, ..., o_t)$ in order to generate the next datapoint of the sequence ($softmax(o_t)$). This newly generated data then gets added as the latest output $o_{t+1}$ \cite{sutskever2011}. Although RNNs are designed mainly for sequential data, CNN based networks can utilize an advantage if the target sequential data is periodic and local 2D features are important to the dataset as in a specific energy consumption profile throughout multiple sample periods \cite{lecun1995}.

\paragraph*{Contribution} After introducing the chosen time based test dataset, the paper shows the conversion of the sequential dataset to 2D and why this conversion intuitively makes sense as well as how much neurons can be economized by this conversion. Afterwards, the dataset is augmented using GANs and evaluated using a process, combining the augmented dataset with a CNN and labelled data. The main contribution of this paper is to check the feasibility of augmenting 1D sequential data from the IoT sector using 2D GANs.

\section{Dataset and Preprocessing}
\label{sec:hecd}

Household energy consumption data is commonly obtained by smart meters to analyse and control smart grid loads. In order to enable non-intrusive load monitoring (NILM), energy consumption data is commonly sampled in a short period, from seconds to several minutes, in combination with energy consumption profiles of household appliances. In addition to enabling the Smart Grid, Smart Home and IoT, data like this, in combination with NILM, can also be used for criminal intents to e.g. analyse if a certain household is currently inhabited or if the inhabitants are on vacation \cite{mclaughlin2011}. Studies based on the dataset used in this paper have aimed to investigate the possibilities of finding security and privacy issues in this kind of smart meter data \cite{ferner2019}.

The dataset consists of 869 Austrian households with their smart meter readings recorded in a 15-min interval. This leads to $37920$ datapoints per household per year. For each household, an equipment list denoting the presence of some consumption-heavy devices (e.g. swimming pool pump, sauna, home cinema) is also available.

As shown by Ferner et al. \cite{ferner2019}, this dataset can be used to analyse if a house has a swimming pool or not, which was chosen as a demo implementation and comparison for this paper. The main difficulty with this dataset is that the data labelled as pool is vastly under-represented, making the training of a classifier CNN complicated, which can be seen in the reported precision value of $0.719$ \cite{ferner2019}. Since this power consumption data is a time based 1D signal showing how much power was utilized each $15$ minutes, a transformation into 2D can be applied. Using the daily periodicity of the signal, it was transformed by taking each period, in this case each day and putting it into a column. Repeating this process for each of the $395$ days converts the 1D signal into a 2D signal that can be interpreted as a heatmap \cite{ferner2019}.

\begin{figure}[h!]
  \includegraphics[width=0.45\textwidth]{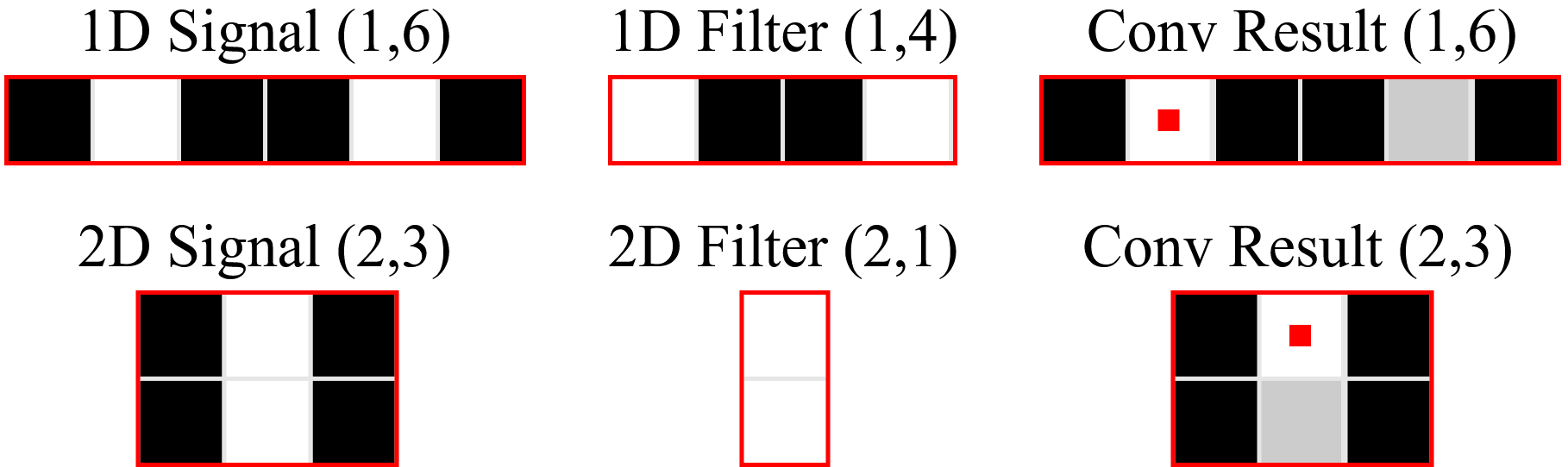}
  \centering
  \caption{Reduction of the convolutional kernel size through a 1D to 2D conversion}
  \label{fig:12convs}
\end{figure}

The transformation into this 2D signal is important since it allows the exploitation of locality through convolutions. The paper aims to investigate if a GAN working with 2D convolutions is capable of generating 1D sequential data, enabling use of sophisticated methods of CNNs like parameter sharing, pooling and striding. Interpreting this 2D signal as pixels in an image, it can intuitively be shown by Fig.~\ref{fig:12convs} that a 1D convolution applied to a 1D signal needs to look at more pixels than a 2D convolution on a 2D signal, further reducing model complexity through convolutional parameter sharing. This outcome results from the way 2D convolutions make use of locality. In more detail, relationships between periods can be exploited along two axis instead of one, resulting in a reduction of needed neurons, in the context of CNNs. If some object to determine via a convolution is placed in the middle of a 2D representation and the distance from the object to the border of the image is called $d$, this reduction is $2d$. If one wants to use these advantages when working with sequential data, a transformation into 2D should be performed.

\section{Sequential Data Generation}
\label{sec:sdg}

There are multiple methods of generating sequential data like conventional data augmentation through a priori knowledge or automated methods like VAEs, RNNs or GANs. GANs have already been used with success in a wide range of image applications as shown by Denton et al. \cite{NIPS2015} and Radford et al. \cite{radford2015}. A GAN is an adversarial network consisting of a generator which tries to generate fake data that is similar to real data (like counterfeit money) as well as a discriminator, which aims to classify real data from fake (like police, checking the real or fake money). This adversarial structure enables training on images, by exploiting established CNN techniques like convolutions in the discriminator and transposed convolutions in the generator \cite{goodfellow2014}.

\begin{figure}[h!]
	\begin{subfigure}[t]{.11\textwidth}
    \centering
    \caption{Non-Pool 1}
    \includegraphics[width=\linewidth]{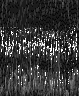}
  	\end{subfigure}
  	\begin{subfigure}[t]{.11\textwidth}
    \centering
    \caption{Non-Pool 2}
    \includegraphics[width=\linewidth]{ChosenGenerated//nPool1.png}
  \end{subfigure}
  \hfill
  \begin{subfigure}[t]{.11\textwidth}
    \centering
    \caption{Pool 1}
    \includegraphics[width=\linewidth]{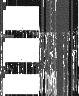}
  \end{subfigure}
  \begin{subfigure}[t]{.11\textwidth}
    \centering
    \caption{Pool 2}
    \includegraphics[width=\linewidth]{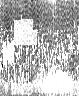}
  \end{subfigure}
  \caption{Real Non-Pool/Pool Heatmaps}
  \label{fig:realHeatmaps}
  \begin{subfigure}[t]{.11\textwidth}
    \centering
    \caption{Non-Pool 1}
    \includegraphics[width=\linewidth]{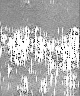}
  \end{subfigure}
  \begin{subfigure}[t]{.11\textwidth}
    \centering
    \caption{Non-Pool 2}
    \includegraphics[width=\linewidth]{ChosenGenerated//gen_nPool1.png}
  \end{subfigure}
  \hfill
  \begin{subfigure}[t]{.11\textwidth}
    \centering
    \caption{Pool 1}
    \includegraphics[width=\linewidth]{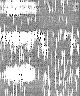}
  \end{subfigure}
  \begin{subfigure}[t]{.11\textwidth}
    \centering
    \caption{Pool 2}
    \includegraphics[width=\linewidth]{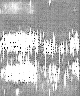}
  \end{subfigure}
  \caption{Generated Non-Pool/Pool Heatmaps}
\label{fig:genHeatmaps}
\end{figure}

Since the chosen energy consumption data has a very high variability with a heavily under-represented class, applying a Deep Convolutional GAN (DCGAN)\footnote{adopted from Naoki Shibuya deep-learning github} did not supply any meaningful results. The main problem during training the DCGAN, vanishing gradients, was successfully countered by switching from the Kullback Leibler Divergence (KLD) to the Earth Mover's Distance (EMD). Additionally, the transfer function of the classifying discriminator neuron is removed and a gradient penalty is introduced, resulting in a WGAN \cite{arjovsky2017_1}. The used WGAN generator consists of $128$ neurons in the input layer (times the input dimensions) and $3$ transposed convolutional layers with $64$, $32$ and $1$ $5\times5$ kernel respectively. Batch norm, leaky ReLUs as well as padding is applied in the generator. The critic consists of $3$ convolutional layers with $32$, $64$ and $128$ convolutions each with $5\times5$ kernels. $2\times2$ Striding as well as padding is applied with leaky ReLUs as transfer functions. After these convolutional layers, the network flattens, followed by $2$ fully connected layers with the first consisting of $1024$ neurons and the second, being the classifier layer, with only $1$ neuron. This layer outputs a score on how good of a fake the generated image is. Adam is chosen as an optimizer with a starting learning rate of $0.0001$ and a learning rate decay of $0.5$. The chosen batch size is $4$ and each GAN was trained for $220$ epochs with a gradient penalty of $10$. After training each GAN, it is used to generate $5000$ images labelled as either pool or non-pool. As Fig. \ref{fig:realHeatmaps} and Fig. \ref{fig:genHeatmaps} show, this WGAN, adopted from the Keras WGAN implementation with updated input dimensions, was subjectively able to learn how to generate household energy consumption heatmaps. Since two WGANs were trained, one with only swimming pool data and one without swimming pool data, two generators resulted from this experiment.

In order to quantitatively evaluate these generators additionally to a subjective analysis, an evaluation workflow was designed. The target was to train a generator with only a subset of all data and then use only the output of this generator to train the classifier network, from \cite{ferner2019}. Therefore, the classifier CNN is trained only with fake data, and evaluated against real data. The assumption is that, if this is successful (successful as in having an f-score $> 0.5$) the generator learned to generate some important representations automatically, resulting in a useful training process for classifier networks like CNNs.

\section{Results and Discussion}
\label{sec:rad}

Fig. \ref{fig:genHeatmaps} shows selected outputs from one of the in Section \ref{sec:sdg} introduced generators. Disregarding the greyish background, the generated household energy consumption heatmaps are very similar to the real data. The generated pool data shows that, about half the time, the usual daily information (noise) is lost but other than that it successfully generates noisy, pool like structures (filled rectangles) in the beginning half of the data (swimming seasons). This seems to suggest that, even with the low amount of training data ($58$ pools), the generator was able to learn the representation of multiple, rectangular shapes. Qualitatively, this suggests that generating household energy consumption data with these kinds of noisy data and simple rectangular features is possible.

\begin{figure}[h!]
  \includegraphics[width=0.25\textwidth]{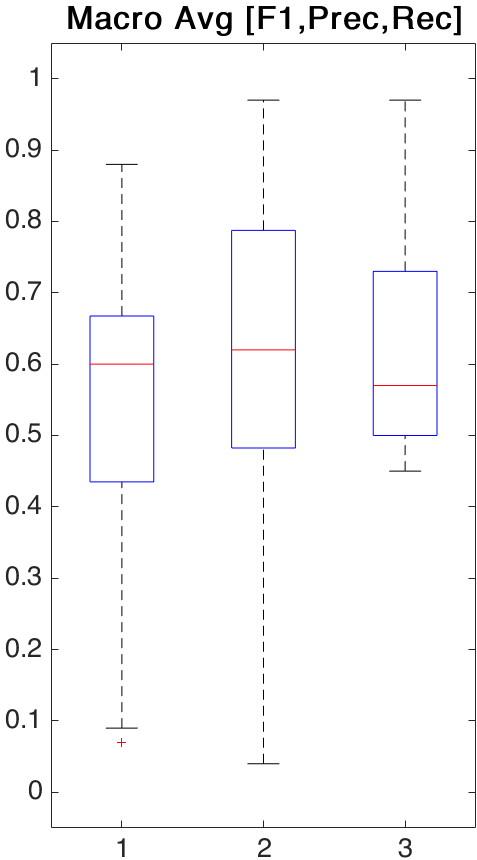}
  \centering
  \caption{Augmented (GAN) classification result}
  \label{fig:boxplot}
\end{figure}

This statement is reinforced by the classification results which can be seen in Fig. \ref{fig:boxplot}. These classification results stem from the in section \ref{sec:sdg} described evaluation process. In detail it is the precision, recall and f-score of a CNN trained by GAN generated data, evaluated against real data. The boxplot shows the macro average over these performance measures of class 0 (without pools) and class 1 (with pools). Since the pool class is heavily under-represented, accuracy should not be used as an evaluation metric. The evaluation shows almost perfect classification for the non-pool class ($0.95$ - $1$ f-score) and about $0.31$ f-score for class $1$. It should be noted that a perfect f-score for class $1$ is impossible since the pool pump feature classes are faulty in some cases and do not appear in every pool dataset. The macro average over class $0$ and $1$ results in a F-score of about $0.6$, precision of $0.62$ and recall of about $0.57$. These values are obtained from 64 independent trials and show that the CNN was able to learn some representations from only generated data. However, the variation of classification scores is substantial.

\begin{table}[h!]
\centering
\begin{tabular}{llllllllll}
\multicolumn{3}{l}{\textbf{Macro Average}} \\
\textbf{F1} & \textbf{Prec} & \textbf{Rec} \\
0.88        & 0.84          & 0.93         \\
0.83        & 0.87          & 0.79         \\
0.82        & 0.91          & 0.77         \\
0.82        & 0.91          & 0.77         \\
0.81        & 0.78          & 0.84        
\end{tabular}
\caption{Top five CNN results}
\label{table:bestGens}
\end{table}

Table \ref{table:bestGens} shows the top 5 CNN classification scores. These generators seem to have learned their respective representations very well. However, due to the high variance in the results, it is still inconclusive if these results stem from the random training through back propagation or possible sampling. 

Using GANs for sequential data generation seems unintuitive since RNNs, especially, LSTMs are designed with sequential data in mind \cite{sutskever2011}. However, as shown in this paper, in some cases, sequential data can be transformed into 2D by exploiting periodic behaviour. In this 2D space, GANs can be used to exploit locality through convolutions and apply further CNN methods like pooling and striding. In these cases, GANs can be used to generate data from noise with a fixed output size, trained from a dataset with distinct representations.

\section{Conclusion}
\label{sec:con}

This paper investigates the possibility of augmenting 1D sequential data using GANs, which is especially interesting on data with a long gathering time (through rare anomaly occasions or a long sequence duration) or on sensitive data with privacy concerns. In order to do so, the paper discusses standard data augmentation methods as well as a set of related work in RNNs and VAEs. After that, the demo dataset from the IoT sector as well as GANs are introduced and data augmentation using GANs is performed. The evaluation of this process shows success in a qualitative as well as quantitative way. However, there is still a huge variation in performance measures, probably due to the small experiment size of $64$ independent trials, that will be increased in future experiments using openly accessible data.

One further step is to evaluate the introduced system against a generative LSTM implementation by comparing their classification scores based on the same input data. Another further step would be to convert the sequential 1D data into 2D, followed by obtaining the local information and converting it back into a 1D space. One example on how to reconvert the data to 1D could be by applying pseudo-hilbert curves \cite{2008effect} or line scans of the whole image, separated into local fields. This would theoretically enable a combination of the advantages of CNNs (locality) and RNNs (sequential relations) by applying this preprocessing to a generative LSTM.

In conclusion, the results of this paper suggest that GANs can be used as a Data Augmentation tool in the ML workflow even if working with sequential data. This leads to shorter data acquisition phases, in e.g. sensor based ML projects for the IoT, and as a possible help to approach the small data challenge. A further investigation of the introduced methods and improvements for the IoT sector is advised.

%\vfill\pagebreak
\bibliographystyle{IEEEbib}
\bibliography{sequential}

\end{document}